\pdfoutput=1
\documentclass[11pt]{article}

\usepackage[]{naacl2021}
\usepackage{times}
\usepackage{latexsym}

\usepackage{times}
\usepackage{latexsym}
\usepackage{url}
\usepackage[T1]{fontenc}
\usepackage[utf8]{inputenc}

\usepackage{microtype}

\usepackage{graphicx}

\usepackage{CJK,algorithm,algorithmic,amssymb,amsmath,array,epsfig,graphics,multirow,array,float,subfigure,verbatim,epstopdf}
\usepackage{enumitem}
\usepackage{color,soul}
\usepackage{hhline}
\usepackage{multirow}
\usepackage{tabularx}
\usepackage{bm}
\usepackage{seqsplit}
\usepackage{multirow, makecell}

\usepackage{hhline}

\usepackage{microtype}
\usepackage{booktabs}

\usepackage{xcolor}

\usepackage{xparse}

\title{COVID-19 Literature Knowledge Graph Construction and Drug Repurposing Report Generation}

\author{
Qingyun Wang\textsuperscript{\textnormal{1}}, 
Manling Li\textsuperscript{\textnormal{1}}, 
Xuan Wang\textsuperscript{\textnormal{1}}, 
Nikolaus Parulian\textsuperscript{\textnormal{1}}, 
Guangxing Han\textsuperscript{\textnormal{2}}, 
\\
\textbf{Jiawei Ma\textsuperscript{\textnormal{2}}, }
\textbf{Jingxuan Tu\textsuperscript{\textnormal{3}}, }
\textbf{Ying Lin\textsuperscript{\textnormal{1}},}
\textbf{Haoran Zhang\textsuperscript{\textnormal{1}}, }
\textbf{Weili Liu\textsuperscript{\textnormal{1}}, } 
\textbf{Aabhas Chauhan\textsuperscript{\textnormal{1}}, }
\\
\textbf{Yingjun Guan\textsuperscript{\textnormal{1}}, }
\textbf{Bangzheng Li\textsuperscript{\textnormal{1}},} 
\textbf{Ruisong Li\textsuperscript{\textnormal{1}}, }
\textbf{Xiangchen Song\textsuperscript{\textnormal{1}},}
\textbf{Yi R. Fung\textsuperscript{\textnormal{1}}, }
\textbf{Heng Ji\textsuperscript{\textnormal{1}}, }
\\
\textbf{Jiawei Han\textsuperscript{\textnormal{1}}, }
\textbf{Shih-Fu Chang\textsuperscript{\textnormal{2}}, }
\textbf{James Pustejovsky\textsuperscript{\textnormal{3}}, }
\textbf{Jasmine Rah\textsuperscript{\textnormal{4}}, }
\textbf{David Liem\textsuperscript{\textnormal{5}}, }
\\
\textbf{Ahmed Elsayed\textsuperscript{\textnormal{6}}, }
\textbf{Martha Palmer\textsuperscript{\textnormal{6}}, }
\textbf{Clare Voss\textsuperscript{\textnormal{7}}, }
\textbf{Cynthia Schneider\textsuperscript{\textnormal{8}}, }
\textbf{Boyan Onyshkevych\textsuperscript{\textnormal{9}}}
\\
  \textsuperscript{1}University of Illinois at Urbana-Champaign \textsuperscript{2}Columbia University
  \textsuperscript{3}Brandeis University \\
  \textsuperscript{4}University of Washington
  \textsuperscript{5}University of California, Los Angeles
  \textsuperscript{6}Colorado University \\
  \textsuperscript{7}Army Research Lab
  \textsuperscript{8}QS2
  \textsuperscript{9}Department of Defense
  \\
  \texttt{hengji@illinois.edu}, 
  \texttt{hanj@illinois.edu},
  \texttt{sc250@columbia.edu}
}

\date{}

\begin{document}
\maketitle

\begin{abstract}
    To combat COVID-19, both clinicians and scientists need to digest vast amounts of relevant biomedical knowledge in scientific literature to understand the disease mechanism and related biological functions. We have developed a novel and comprehensive knowledge discovery framework, \textbf{COVID-KG} to 
    extract fine-grained multimedia knowledge elements (entities and their visual chemical structures, relations and events) from scientific literature.  We then exploit the constructed multimedia knowledge graphs (KGs) for question answering and report generation, using drug repurposing as a case study. Our framework also provides detailed contextual sentences, subfigures, and knowledge subgraphs as evidence. All of the data, KGs, reports\footnote{Demo video: \url{http://159.89.180.81/demo/covid/Covid-KG\_DemoVideo.mp4}}, resources, and shared services are publicly available\footnote{Project website: \url{http://blender.cs.illinois.edu/covid19/}}.
    \end{abstract}
\section{Introduction}
Practical progress at combating COVID-19 highly depends on effective search, discovery, assessment and extension of scientific research results.
However, clinicians and scientists are facing two unique barriers on digesting these research papers.

\begin{figure}[!hbt]
\centering
\includegraphics[width=\linewidth]{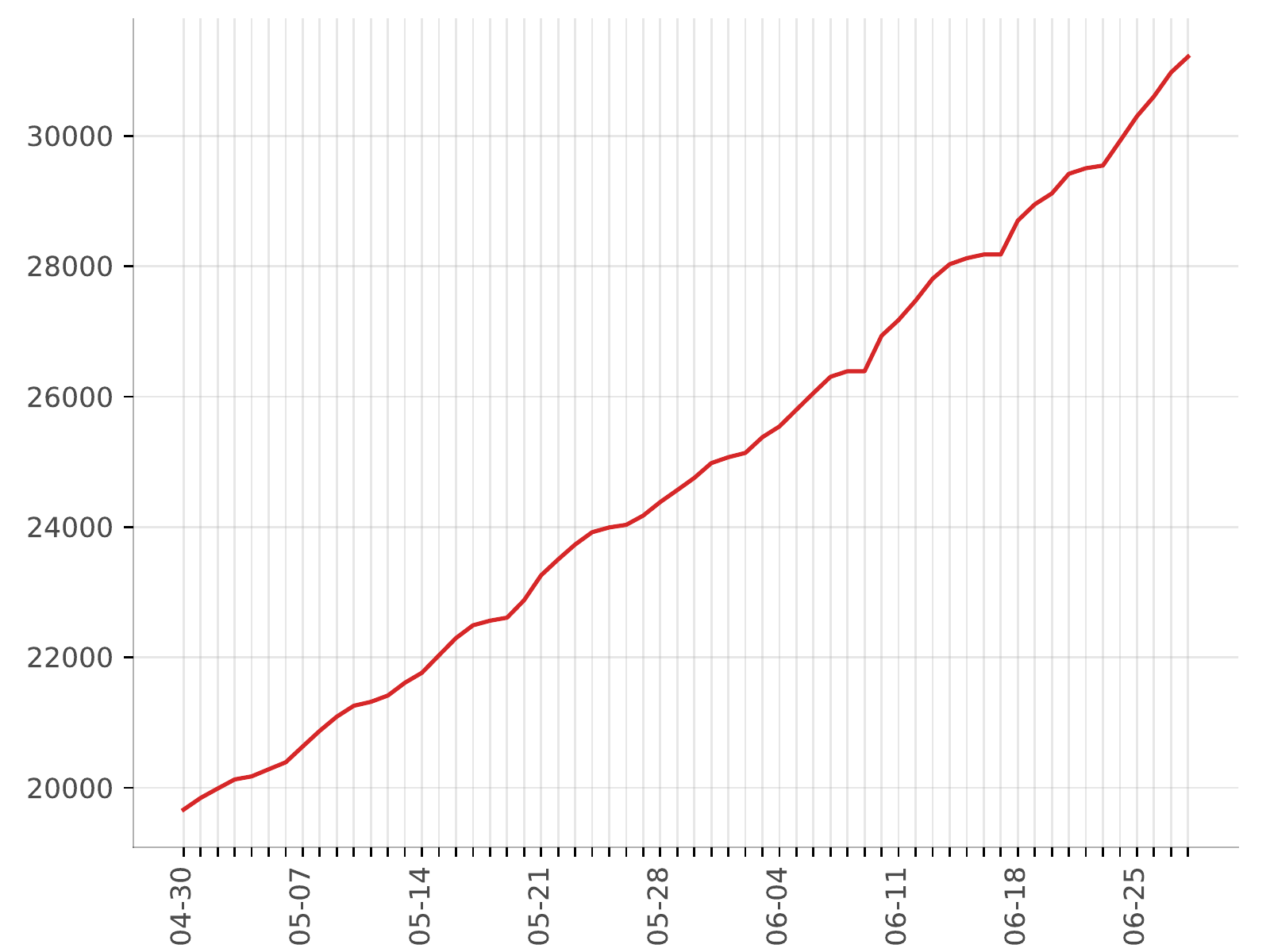}
\caption{The Growing Number of COVID-19 Papers at PubMed}
\label{fig:covid_curve}
\end{figure}

\begin{figure*}
    \centering
    \includegraphics[width=\linewidth]{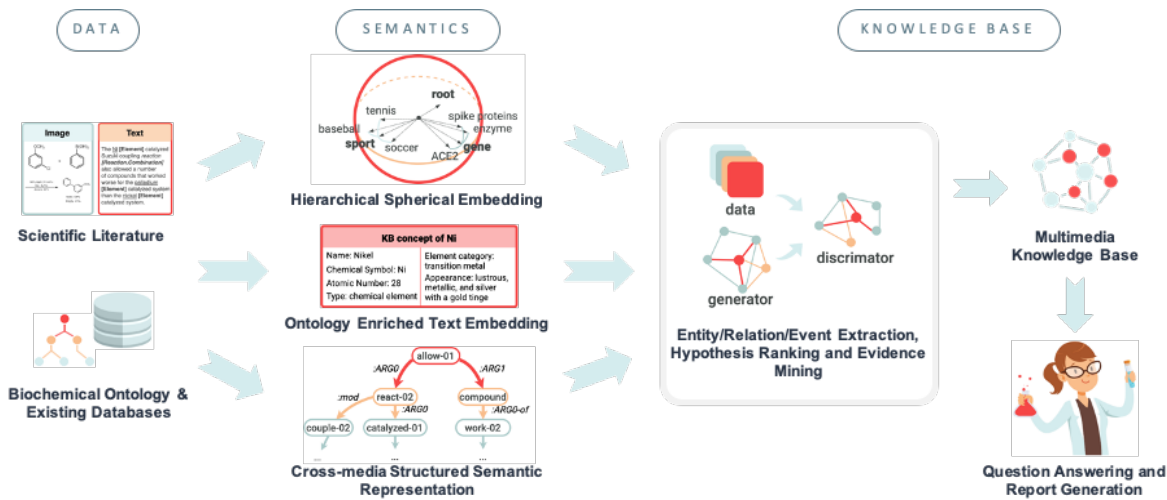}
    \caption{COVID-KG Overview: From Data to Semantics to Knowledge }
    \label{fig:overview}
\end{figure*}

The first challenge is \emph{quantity}. Such a bottleneck in knowledge access is exacerbated during a pandemic when  increased investment in relevant research  leads to even faster growth of literature than usual.
For example, as of April 28, 2020, at PubMed\footnote{https://www.ncbi.nlm.nih.gov/pubmed/} there were 19,443 papers related to coronavirus; as of June 13, 2020, there were 140K+ related papers, \emph{nearly 2.7K new papers per day} (see Figure~\ref{fig:covid_curve}). 
The resulting knowledge bottleneck contributes significant delays in the development of vaccines and drugs for COVID-19. More intelligent knowledge discovery technologies need to be developed to enable researchers to more quickly and accurately access and digest relevant knowledge from the literature.

The second challenge is \emph{quality}. 
Many research results about coronavirus from different research labs and sources are redundant, complementary, or even conflicting with each other, while some false information has been promoted in both formal publication venues as well as social media platforms such as Twitter. As a result, some of public policy responses to the virus, and public perception of it, have been based on misleading, and at times erroneous, claims. The relative isolation of these knowledge resources makes it hard, if not impossible, for researchers to connect the dots that exist in separate resources to gain new insights. 

Let us consider drug repurposing as a case study.\footnote{This is a \emph{pre-clinical phase of biomedical research} to discover new uses of existing, approved drugs that have already been tested in humans and so detailed information is available on their pharmacology, formulation and potential toxicity.} Besides the long process of clinical trial and biomedical experiments, another major cause of the lengthy discovery phase is the complexity of the problem involved and the difficulty in drug discovery in general. The current clinical trials for drug repurposing rely mainly on reported symptoms in considering drugs that can treat diseases with similar symptoms. However, there are too many drug candidates and too much misinformation published in multiple sources. The clinicians and scientists thus urgently need assistance in obtaining a reliable ranked list of drugs with detailed 
evidence,  
and also in gaining new insights into the underlying molecular cellular mechanisms on COVID-19 and the pre-existing conditions that may affect the mortality and severity of this disease.

To tackle these two challenges we propose a new framework, \textbf{COVID-KG}, to accelerate scientific discovery and build a bridge between the research scientists making use of our framework and clinicians who will ultimately conduct the tests, as illustrated in Figure~\ref{fig:overview}.   
\textbf{COVID-KG} starts by reading existing papers to build multimedia knowledge graphs (KGs), in which nodes are entities/concepts and edges represent relations and events involving these entities, as extracted from both text and images. 
Given the KGs enriched with path ranking and evidence mining, \textbf{COVID-KG} answers natural language questions effectively. With drug repurposing as a case study, we focus on 11 typical questions that human experts pose and integrate our techniques to generate a comprehensive report for each candidate drug.

\section{Multimedia Knowledge Graph Construction}
\label{sec:kgconstruction}

\begin{figure}[!hbt]
\centering
\includegraphics[width=\linewidth]{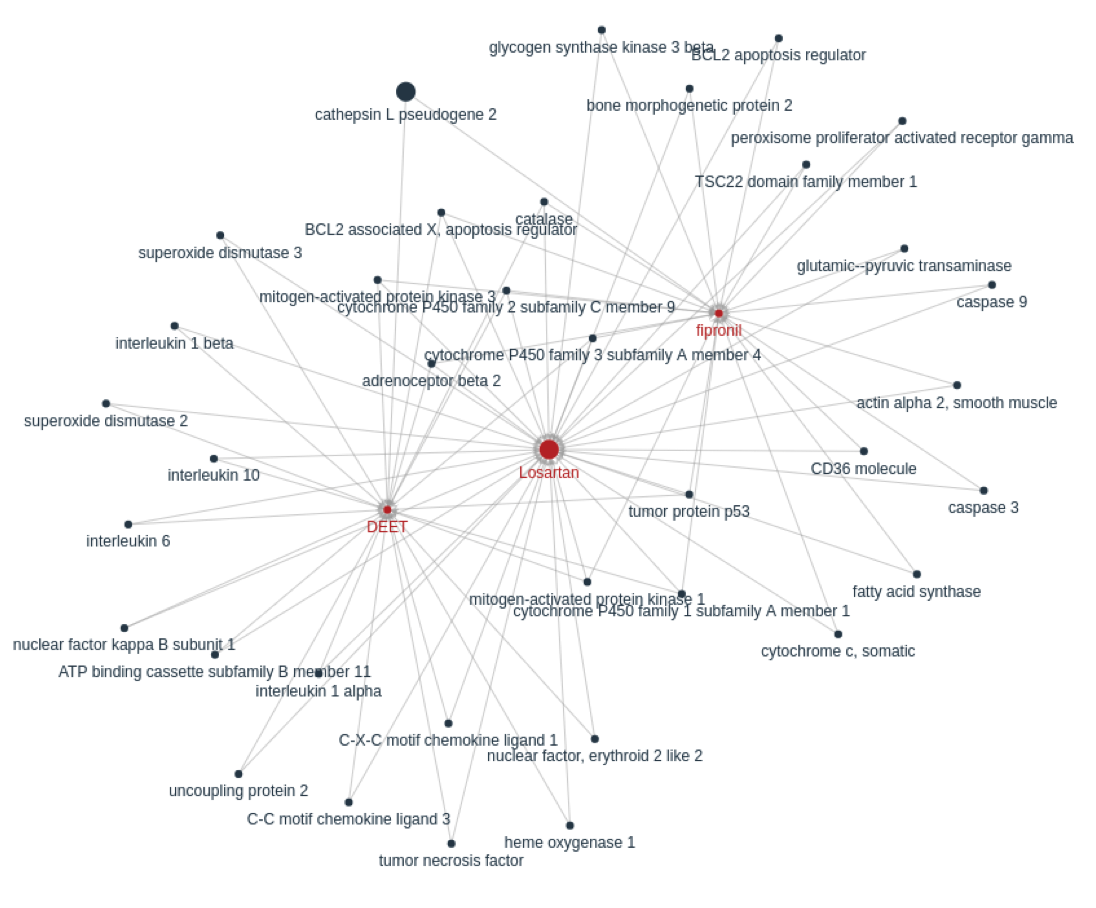}
\caption{Constructed KG Connecting Losartan (candidate drug in COVID-19) and cathepsin L pseudogene 2 (gene related to coronavirus),  where red nodes represent chemicals, grey nodes represent genes, and edges represent gene-chemical relations.}
\label{fig:kg}
\end{figure}

\subsection{Coarse-grained Text Knowledge Extraction}

Our coarse-grained Information Extraction (IE) system consists of three components: (1) coarse-grained entity extraction~\cite{wang-etal-2019-paperrobot} and entity linking~\cite{Zheng2014} for four entity types: \emph{Gene nodes, Disease nodes, Chemical nodes, and Organism.} We follow the entity ontology defined in the Comparative Toxicogenomics Database (CTD)~\cite{davis2016comparative},  and obtain a Medical Subject Headings (MeSH) Unique ID for each mention. (2) Based on the MeSH Unique IDs, we further link all entities to the CTD and extract 133 subtypes of relations such as 
\emph{Gene–Chemical–Interaction Relationships, Chemical–Disease Associations, Gene–Disease Associations, Chemical–GO Enrichment Associations and Chemical–Pathway Enrichment Associations.} (3) Event extraction~\cite{li-etal-2019-biomedical}: we extract 13 Event types and the roles of entities involved in these events as defined in \cite{nedellec-etal-2013-overview}, including \emph{Gene expression, Transcription, Localization, Protein catabolism, Binding, Protein modification, Phosphorylation, Ubiquitination, Acetylation, Deacetylation, Regulation, Positive regulation, and Negative regulation.} Figure~\ref{fig:kg} shows an example of the constructed KG from multiple papers. Experiments on 186 documents with 12,916 sentences manually annotated by domain experts show that our method achieves 83.6\% F-score on node extraction and 78.1\% F-score on link extraction.

\subsection{Fine-grained Text Entity Extraction}

\begin{figure}[!hbt]
\centering
\includegraphics[width=0.5\textwidth]{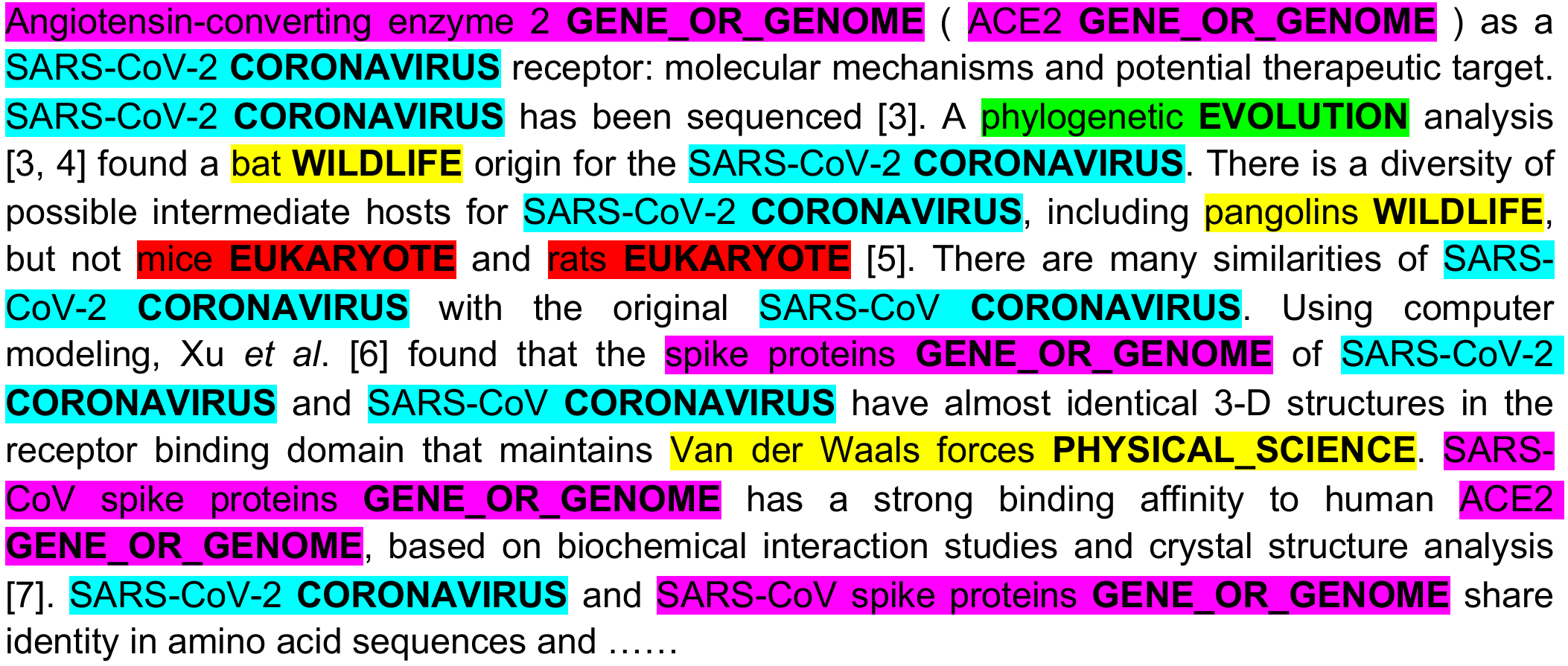}
\caption{Example of Fine-grained Entity Extraction}
\label{fig:cord-ner}
\end{figure}

However, questions from experts often involve fine-grained knowledge elements, such as \emph{“Which \textbf{amino acids} in glycoprotein
are most related to Glycan (CHEMICAL)?”.} 
To answer these questions, 
we apply our
fine-grained entity extraction system CORD-NER 
\cite{wang2020comprehensive} to extract 75 types of entities to enrich the KG, including 
many COVID-19 specific new entity types 
(e.g., \emph{coronaviruses, viral proteins, evolution, materials, substrates and immune responses}).
CORD-NER relies on distantly- and weakly-supervised methods \cite{wang2019distantly, shang-etal-2018-learning}, with no need for expensive human annotation. 
Its entity annotation quality surpasses SciSpacy (up to 93.95\% F-score, over 10\% higher on the F1 score based on a sample set of documents), a fully supervised BioNER tool. 
See Figure \ref{fig:cord-ner} for results on part of a COVID-19 paper \cite{zhang2020angiotensin}.

\subsection{Image Processing and Cross-media Entity Grounding}
\begin{figure}[t]
    \centering
    \includegraphics[width=\linewidth]{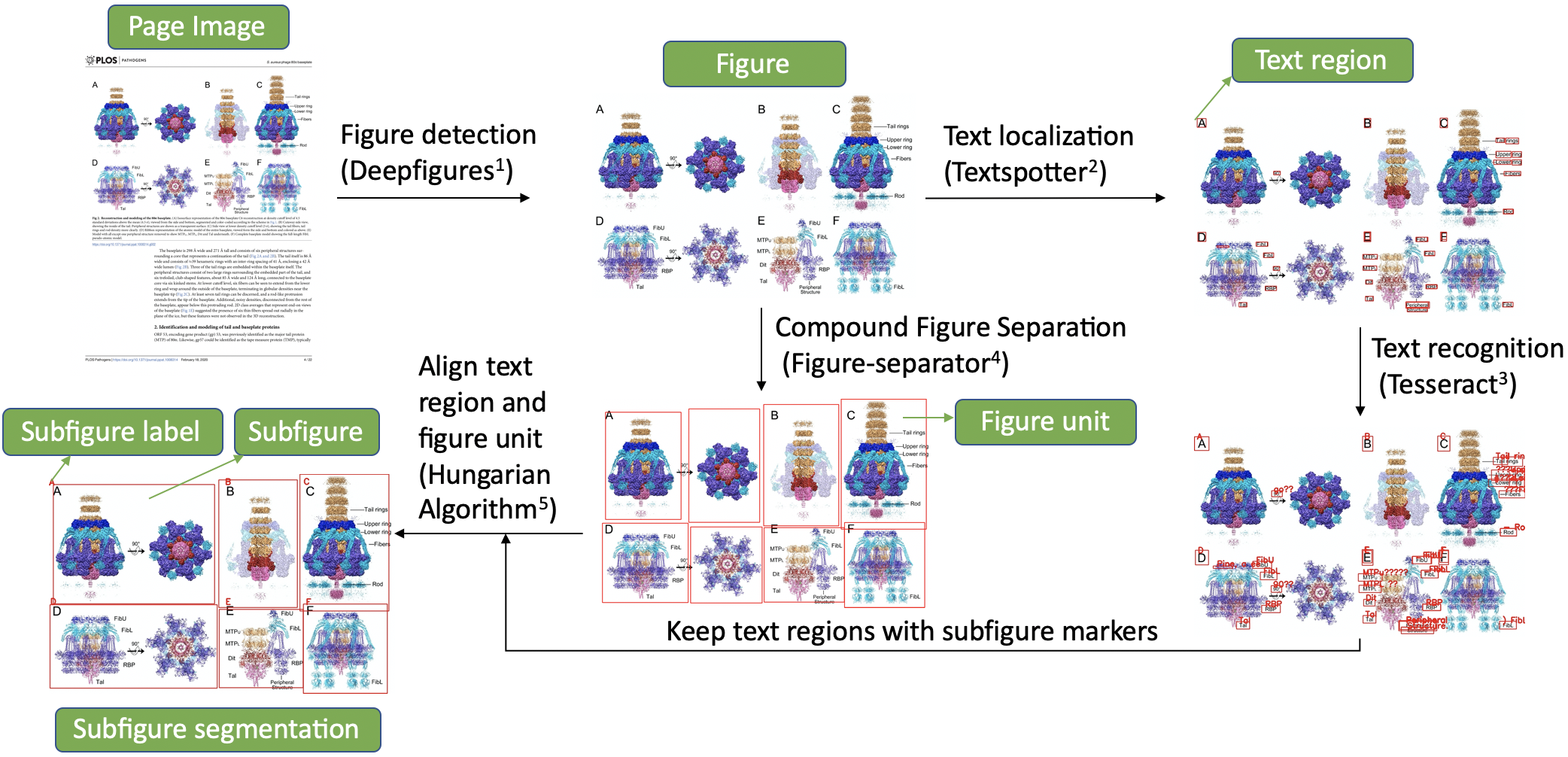}
    \caption{System Pipeline for Automatic Figure Extraction and Subfigure Segmentation. The figure image shown here is from \cite{kizziah2020structure}}
    \label{fig:pipeline_subfigure_analysis}
\end{figure}

Figures in biomedical papers may contain different types of visual information, for example, displaying molecular structures, microscopic images, dosage response curves, relational diagrams, and other uniquely visual content.  We have developed a visual IE subsystem to extract the visual information from figures to enrich the KG. We start by designing a pipeline and automatic tools shown in Figure \ref{fig:pipeline_subfigure_analysis} to extract figures from papers in the CORD-19 dataset and segment figures into nearly half a million isolated subfigures. In the end, we perform cross-modal entity grounding, i.e., associating visual objects identified in these subfigures with entities mentioned in their captions or referring text. To start, since most figures are embedded as part of PDF files,
we run Deepfigures \cite{siegel2018extracting} to automatically detect and extract figures from each PDF document. Then each figure is associated with text in its caption or referring context (main body text referring to the figure). In this way, a figure can be attached, at a coarse level, to a KG entity if that entity is mentioned in the associated text.

\begin{figure}
    \centering
    \includegraphics[width=\linewidth]{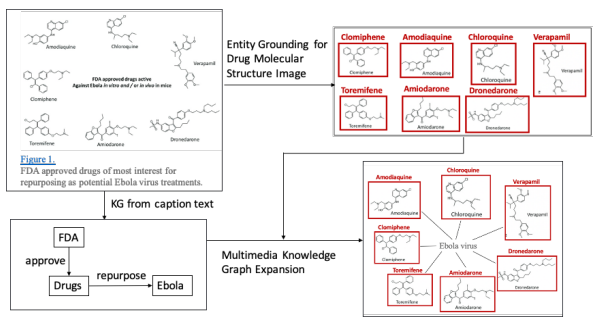}
    \caption{Expanding KG through Subfigure Segmentation and Cross-modal Entity Grounding. The figure image shown here is from \cite{ekins2015fda}}
    \label{fig:multimedia_kg_construction}
\end{figure}

To further delineate semantic and visual information contained within each subfigure, we have developed a pipeline to segment individual subfigures and then align each subfigure with its corresponding subcaption. We run Figure-separator \cite{tsutsui2017data} to detect and separate all non-overlapping image regions. On occasion, subfigures within a figure may also be marked with alphabetical letters (e.g., A, B, C, etc). We use deep neural networks \cite{zhou2017east} to detect text within figures and then apply OCR tools \cite{smith2007overview} to automatically recognize text content within each figure. To identify \emph{subfigure marker text} and \emph{text labels} for analyzing figure content, we rely on the distance between text labels and subfigures to locate subfigure text markers. Location information of such text markers can also be used to merge multiple image regions into a single subfigure. At the end, each subfigure is segmented, and associated with its corresponding subcaption and referring context. The segmented subfigures and associated text labels provide rich information that can expand the KG constructed from text captions. For example, as shown in Figure \ref{fig:multimedia_kg_construction}, we apply a classifier to detect subfigures  containing molecular structures. Then by linking the specific drug names extracted from  within-figure text to corresponding drug entities in the coarse KG constructed from the caption text, an expanded cross-modal KG can be constructed that then links images with specific molecular structures to their drug entities in the KG.

\subsection{Knowledge Graph Semantic Visualization}

In order to enhance the exploration and discovery of the information mined from the COVID-19 literature through the algorithms discussed in previous sections, we create semantic visualizations over large 
complex networks of biomedical relations using the techniques proposed by \citet{semviz}. Semantic visualization allows for visualization of user-defined subsets of these relations interactively through semantically typed tag clouds and heat maps. This allows researchers to get a global view of selected relation subtypes drawn from hundreds or thousands of papers at a single glance. This in turn allows for the ready identification of novel relations that would typically be missed by directed keyword searches or simple unigram word cloud or heatmap displays.\footnote{\url{https://www.semviz.org/}}

We first build a data index from the knowledge elements in the constructed KGs, 
and then create a Kibana dashboard\footnote{\url{https://github.com/elastic/kibana}} out of the generated data indices. Each Kibana dashboard has a collection of visualizations that are designed to interact with each other. Dashboards are implemented as web applications. The navigation of a dashboard is mainly through clicking and searching. By clicking the protein keyword \texttt{EIF2AK2} in the tag cloud named \emph{``Enzyme proteins participating Modification relations''}, a constraint on the type of proteins in modifications is added. Correspondingly, all the other visualizations will be changed.

\begin{figure}[ht]
    \centering
    \includegraphics[width=\linewidth]{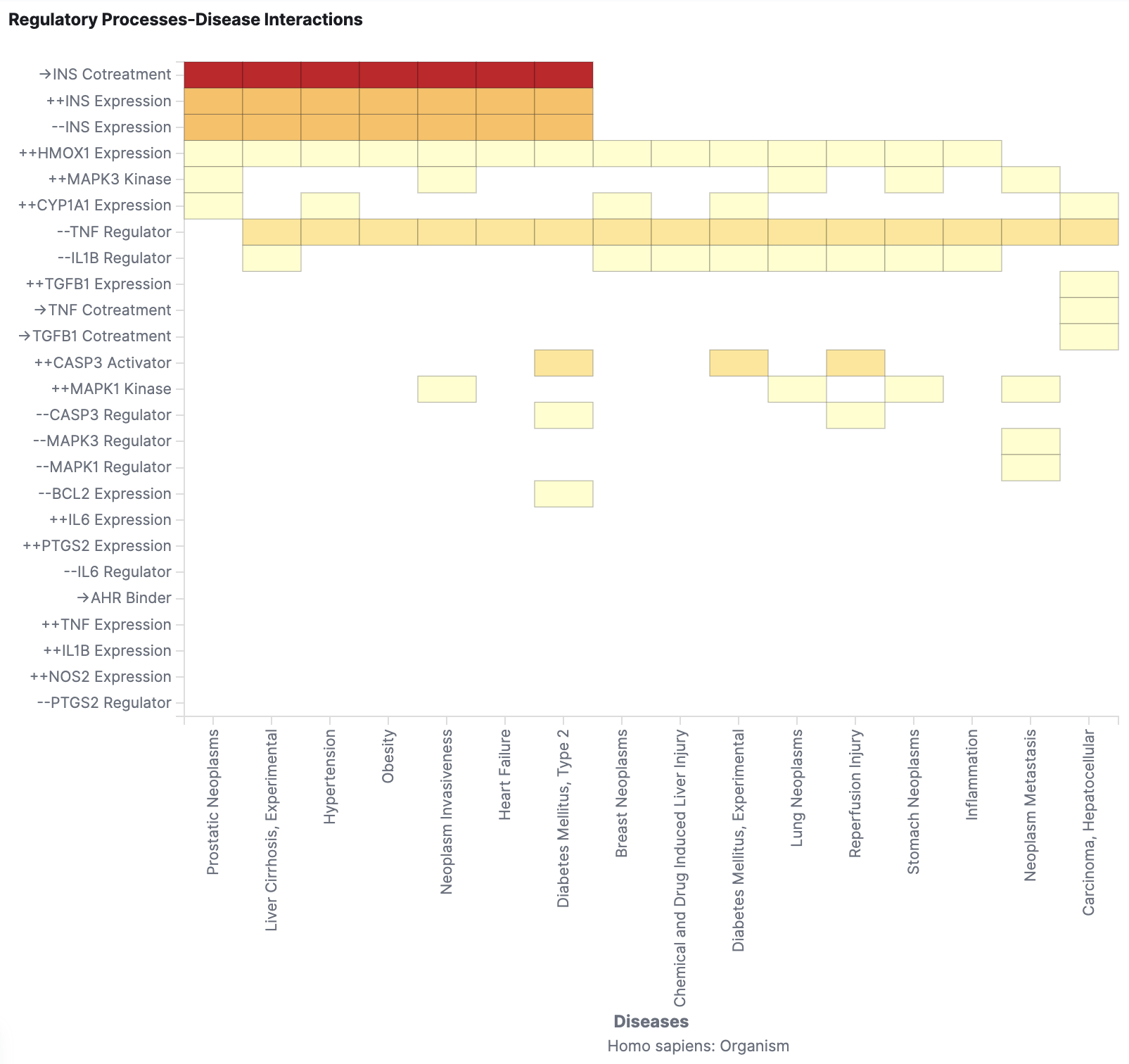}
    \caption{Regulatory Processes-Disease Interactions Heatmap}
    \label{fig:regulatoryHeatmap}
\end{figure}

One unique feature of the semantic visualization is the creation of {\it dense tag clouds} and {\it dense heatmaps}, through a process of parameter reduction over relations, allowing for the visualization of  relation sets as tag clouds and multiple chained relations as heatmaps.  Figure \ref{fig:regulatoryHeatmap} illustrates such a dense heatmap that contains relations between proteins and implicated diseases (e.g., \emph{``those proteins that are down-regulators of TNF which are implicated in obesity''}), along with their type information\footnote{We use the following symbols to indicate the ``action'' involved in each protein: ``$++$'' = increase, ``$--$'' = decrease, ``$\rightarrow$'' = affect.}. 

\section{Knowledge-driven Question Answering}

In contrast to most  current question-answering (QA) methods which target  single documents, we have developed a QA component based on a combination of KG matching and distributional semantic matching across documents. 
We build KG indexing and searching functions to facilitate effective and efficient search when users pose their questions. We also support extended semantic matching from the constructed KGs and related texts by accepting multi-hop queries. 

A common category of queries is about the connections between two entities. 
Given two entities in a query, we generate a subgraph covering salient paths between them to show how they are connected through other entities. \figurename~\ref{fig:kg} is an example subgraph summarizing the connections between \textit{Losartan} and \textit{cathepsin L pseudogene 2}. The paths are generated by traversing the constructed KG, and are ranked by 
the number of papers covering the knowledge elements in each path in the KG. 
Each edge is assigned a salience score by aggregating the scores of paths passing through it. 
In addition to knowledge elements, we also present related sentences and source information as evidence. 
We use BioBert~\cite{lee2020biobert}, a pre-trained language model to represent each sentence along with its left and right neighboring sentences as local contexts. 
Using the same architecture computed on all respective sentences and the user query, we aggregate the sequence embedding layer, the last hidden layer in the BERT architecture with average pooling \cite{reimers-gurevych-2019-sentence}. We use the similarity between the embedding representations of each sentence and each query to identify and extract the most relevant sentences as evidence.

Another common category of queries includes entity types, rather than entity instances, and requires extracting evidence sentences based on type or pattern matching. We have developed \textsc{EvidenceMiner} \cite{wang2020evidenceminer, wang2020automatic}, a web-based system that allows for the user’s query as a natural language statement or an inquiry about a relationship at the meta-symbol level (e.g., CHEMICAL, PROTEIN) and then automatically retrieves textual evidence from a background corpora of COVID-19.

\section{A case study on Drug Repurposing Report Generation}

\subsection{Task and Data}
A human-written report about drug repurposing usually answers the following typical questions. 

\begin{enumerate}[itemsep=-5pt,topsep=0pt]
\item Current indication: what is the drug class? What is it currently approved to treat?
\item Molecular structure (symbols desired,  but a pointer to a reference is also useful)
\item Mechanism of action i.e., inhibits viral entry,  replication,  etc. (w/ a pointer to data)
\item Was the drug identified by manual or computation screen?
\item Who is studying the drug? (Source/lab name)
\item In vitro Data available (cell line used,  assays run,  viral strain used,  cytopathic effects,  toxicity,  LD50,  dosage response curve,  etc.)
\item Animal Data Available (what animal model,  LD50,  dosage response curve,  etc.)
\item Clinical trials on going (what phase,  facility,  target population,  dosing,  intervention etc.)
\item Funding source
\item Has the drug shown evidence of systemic toxicity?
\item List of relevant sources to pull data from.
\end{enumerate}

The answers to questions \#5 and \#11 are extracted based on the meta-data sections of research papers in scientific literature, including the author affiliation and acknowledgement sections. The answers for other questions are all extracted based on the knowledge graphs constructed and knowledge-driven question-answering method described above. 

As in our case studies, DARPA biologists inquired about three drugs, Benazepril, Losartan, and Amodiaquine, and their links to COVID-19 related chemicals/genes as shown in Figure~\ref{fig:chemical}:

\begin{figure}[!hbt]
    \centering
    \includegraphics[width=\linewidth]{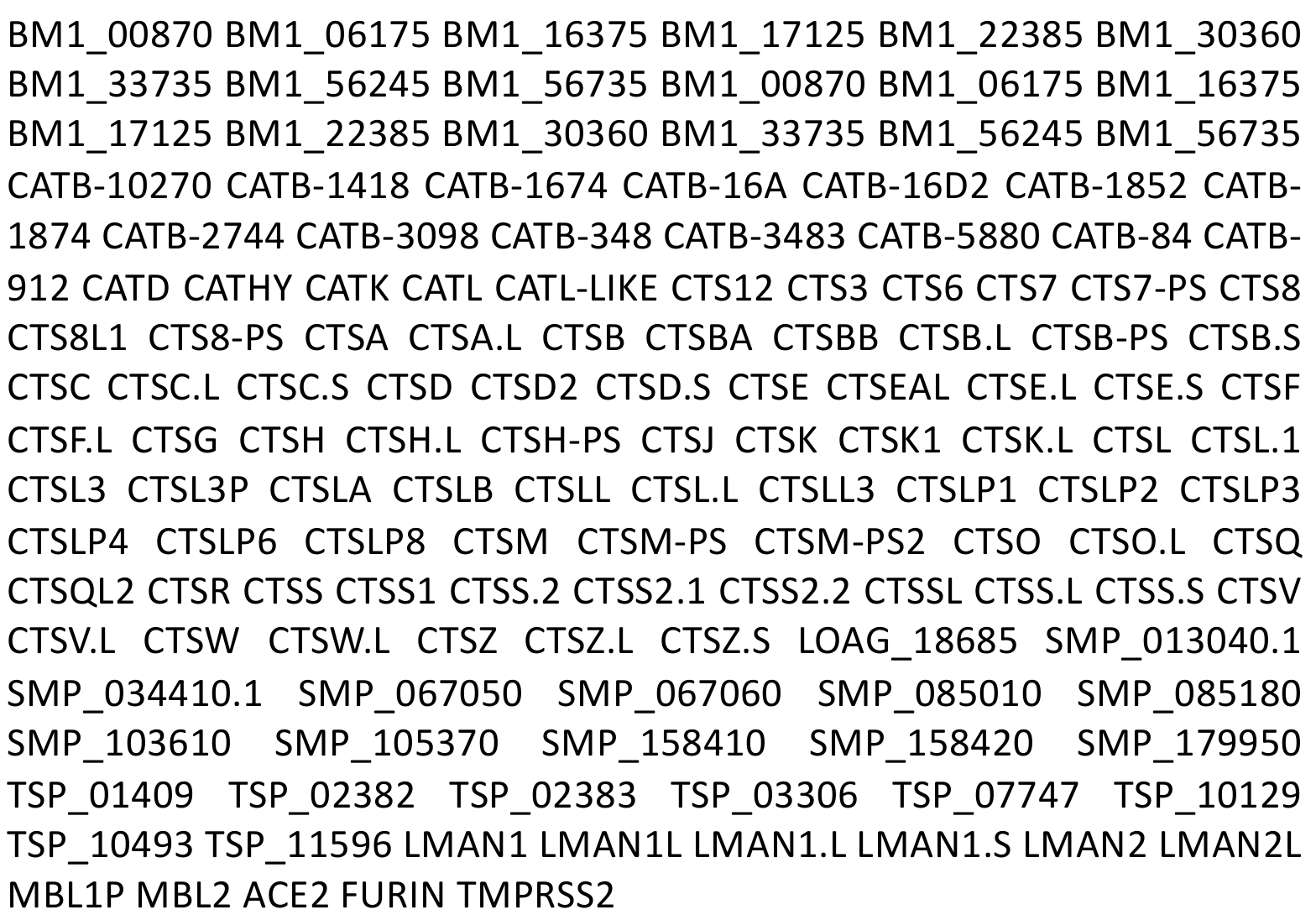}
    \caption{COVID-19 related chemicals/genes.}
    \vspace{-2pt}
    \label{fig:chemical}
\end{figure}

Our KG results for many other drugs are visualized at our website\footnote{\url{http://blender.cs.illinois.edu/covid19/visualization.html}}. 
We download new COVID-19 papers from three Application Programming Interfaces (APIs): NCBI PMC API, NCBI Pubtator API and CORD-19 archive. We provide incremental updates including new papers, removed papers and updated papers, and their metadata information at our website\footnote{\url{http://blender.cs.illinois.edu/covid19/}}.

\subsection{Results}
As of June 14, 2020 we collected 140K papers. We selected 25,534 peer-reviewed papers and constructed the KG that  
includes 7,230 Diseases, 9,123 Chemicals and 50,864 Genes, with 1,725,518 Chemical-Gene links, 5,556,670 Chemical-Disease links, and 77,844,574 Gene-Disease links. The KG has received more than 1,000+ downloads. Our final generated reports\footnote{\url{http://blender.cs.illinois.edu/covid19/DrugRe-purposingReport_V2.0.docx}} are shared publicly. For each question, our framework provides answers along with detailed evidence, knowledge subgraphs and image segmentation and analysis results. Table~\ref{table:good_answers} shows some example answers. 

\begin{figure}[!ht]
    \centering
    \includegraphics[width=0.8\linewidth]{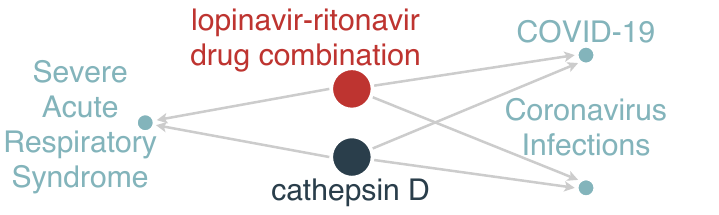}
    \caption{Connections Involving Coronavirus Related Diseases}
    \label{fig:connections_involving_coronavirus_related_diseases}
    \vspace{-6pt}
\end{figure}

Several clinicians and medical school students in our team have manually reviewed the drug repurposing reports for three drugs, and also the KGs connecting 41 drugs and COVID-19 related chemicals/genes. In checking the evidence sentences and reading the original articles, they reported that most of our output is informative and valid. For instance, after the coronavirus enters the cell in the lungs, it can cause a severe disease called Acute Respiratory Distress Syndrome. This condition causes the release of inflammatory molecules in the body named cytokines such as Interleukin-2, Interleukin-6, Tumor Necrosis Factor, and Interleukin-10. We see all of these connections in our results, such as the examples shown in Figure~\ref{fig:kg} and Figure~\ref{fig:connections_involving_coronavirus_related_diseases}. With further checks on these results, the scientists also indicated that many results were worth further investigation. For example, in Figure~\ref{fig:kg} we can see that Lusartan is connected to tumor protein p53 which is related to lung cancer.

\begin{table}[!hbt]
\scriptsize
\centering
\setlength\tabcolsep{2pt}
\setlength\extrarowheight{1pt}
\begin{tabularx}{\linewidth}{|c|l|p{0.67\linewidth}|}
\hline
\textbf{Question} & \multicolumn{2}{c|}{\textbf{Example Answers} }
\\\hline
\multirow{8}{*}{Q1} & 
Drug Class & angiotensin-converting enzyme (ACE) inhibitors
\\\cline{2-3}
& Disease & hypertension
\\\cline{2-3}
& \multirow{5}{*}{Evidence} 
& [PMID:32314699 (PMC7253125)] Past medical history was significant for hypertension,  treated with amlodipine and benazepril,  and chronic back pain. 
\\\cline{3-3}
& \multirow{1}{*}{Sentences} 
& [PMID:32081428 (PMC7092824)] On the other hand, many ACE inhibitors are currently used to treat hypertension and other cardiovascular diseases. Among them are captopril, perindopril, ramipril, lisinopril, benazepril, and moexipril. 
\\\hline
\multirow{8}{*}{Q4} & 
Disease & COVID-19
\\\cline{2-3}
& \multirow{8}{*}{Evidence} 
& [PMID:32081428 (PMC7092824)] By using a molecular docking approach, an earlier study identified N-(2-aminoethyl)-1 aziridine-ethanamine as a novel ACE2 inhibitor that effectively blocks the SARS-CoV RBD-mediated cell fusion. 
\\
& \multirow{1}{*}{Sentences} 
& This has provided a potential candidate and lead compound for further therapeutic drug development. Meanwhile, biochemical and cell-based assays can be established to screen chemical compound libraries to identify novel inhibitors.
\\\hline
\multirow{8}{*}{Q6} & 
Disease & cardiovascular disease
\\\cline{2-3}
& \multirow{5}{*}{Evidence} 
& [PMID:22800722 (PMC7102827)] The in vitro half-maximal inhibitory concentration (IC50) values of food-derived ACE inhibitory peptides are about 1000
\\
& \multirow{1}{*}{Sentences} 
& 
fold higher than that of synthetic captopril but they have higher in vivo activities than would be expected from their in vitro activities.....
\\
% & \multirow{1}{*}{Sentences} 
% & Germinal ACE depends on chloride to a lesser extent compared with the C domain of sACE. Cushman and Cheung reported an optimal in vitro ACE activity of rabbit rung acetone extract in the presence of 300 mM NaCl at pH 8.1-8.3…
% \\
\hline
\multirow{8}{*}{Q8} & 
Disease & COVID-19
\\\cline{2-3}
& \multirow{12}{*}{Evidence} 
& [PMID:32336612 (PMC7167588)] 
Two trials of losartan as additional treatment for SARS-CoV-2 infection in hospitalized (NCT04312009) or not hospitalized (NCT04311177) patients have been announced,  supported by the background of the huge adverse impact of the ACE Angiotensin II AT1 receptor axis over-activity in these patients.
\\\cline{3-3}
& \multirow{1}{*}{Sentences} 
& [PMID:32350632 (PMC7189178)] To address the role of angiotensin in lung injury,  there is an ongoing clinical trial to examine whether losartan treatment affects outcomes in COVID-19 associated ARDS (NCT04312009).
\\\cline{3-3}
& 
& [PMID:32439915 (PMC7242178)] Losartan was also the molecule chosen in two trials recently started in the United States by the University of Minnesota to treat patients with COVID-19 (clinical trials.gov NCT04311177 and NCT 104312009). 
\\
\hline
\end{tabularx}
\caption{Example Answers for Questions in Drug Repurposing Reports}
\label{table:good_answers}
\end{table}

\section{Related Work}

Extensive prior research work has focused on extracting 
biomedical entities~\cite{
zheng2014entity,habibi2017deep,korhonen2017neural,wang2018cross,beltagy-etal-2019-scibert, alsentzer-etal-2019-publicly, 10.1093/nar/gkz389,wang2020comprehensive}, relations~\cite{uzuner20112010,krallinger2011protein,semeval-2013-joint-lexical, 
bui2014novel, peng2016improving, wei2015overview, peng2017cross,
%quirk2016distant,
luo2017bridging,10.1093/nar/gkz389,Libio2019,peng-etal-2019-transfer,peng-etal-2020-empirical}, and events~\cite{ananiadou2010event,van2013large,nedellec-etal-2013-overview,
deleger-etal-2016-overview,10.1093/nar/gkz389,li-etal-2019-biomedical,shafieibavani-etal-2020-global} from biomedical literature, with the most recent work focused on COVID-19 literature~\cite{hope2020scisight,ilievski2020kgtk,wolinski2020visualization,ahamed2020information}.

Most of the recent biomedical QA work~\cite{yang2015learning,yang-etal-2016-learning, chandu-etal-2017-tackling, kraus2017olelo} is driven by the BioASQ initiative \cite{tsatsaronis2015overview}, and many 
live QA systems, including COVIDASK\footnote{\url{https://covidask.korea.ac.kr/}} and  AUEB\footnote{\url{http://cslab241.cs.aueb.gr:5000/}}, and search engines~\cite{kricka2020artificial, esteva2020co, hope2020scisight, taub-tabib-etal-2020-interactive} have been developed. 
Our work is an application and extension of our recently developed multimedia knowledge extraction system for news domain~\cite{li-etal-2020-gaia,li-etal-2020-cross}. Similar to news domain, the knowledge elements extracted from text and images in literature are complementary. Our framework advances state-of-the-art by extending the knowledge elements to more fine-grained types, incorporating image analysis and cross-media knowledge grounding, and KG matching into QA.
\section{Conclusions and Future Work}

We have developed a novel framework, \textbf{COVID-KG}, that 
automatically transforms a massive scientific literature corpus into organized, structured, and actionable KGs, and uses it to answer questions in drug repurposing reporting. 
With \textbf{COVID-KG}, researchers and clinicians are able to obtain informative 
answers from scientific literature, and thus focus on more important hypothesis testing, and prioritize the analysis efforts for candidate exploration directions. In our ongoing work we have created a new ontology that includes 77 entity subtypes and 58 event subtypes, and we are  building a neural IE system following this new ontology. In the future we plan to extend \textbf{COVID-KG} to automate the creation of new hypotheses by predicting new links. 
We will also create a multimedia common semantic space~\cite{li-etal-2020-gaia,li-etal-2020-cross} for literature and apply it to improve cross-media knowledge grounding and inference.

\section*{Ethical Considerations}

\subsection*{Required Workflow for Using Our System} 

\textbf{Human review required.} Our knowledge discovery tool provides investigative leads for pre-clinical research, not final results for clinical use. Currently, biomedical researchers scour the literature to identify candidate drugs, then follow a standard research methodology to investigate their actual utility (involving literature reviews, computer simulations of drug mechanisms and effectiveness, in-vitro studies, cellular in-vivo studies, etc. before moving to clinical studies.). Our tool COVID-KG (and all knowledge discovery tools for biomedical applications) is not meant to be used for direct clinical applications on any human subjects. Rather, our tool aims to highlight unseen relations and patterns in large amounts of scientific textual data that would be too time consuming for manual human effort. Accordingly, the tool would be useful for stakeholders (e.g., biomedical scientists) 
to identify specific drug candidates and molecular targets that are relevant in their biomedical and clinical research aims. Use of our knowledge discovery tool allows the researcher to narrow down the set of candidate drugs to investigate rapidly, but then proceed with the usual sequence of steps before kicking off expensive and time-consuming clinical tests. Failure to follow this sequence of events, and use of the system without the required human review, could lead to misguided experimental design wasting time and resources.

\textbf{Check evidence and source before use our system results.} In addition, our tool provides source and rich evidence sentences for each node and link in the KG. To curtail potential harms caused by extraction errors, users of the knowledge graphs should double check the source information and verify the accuracy of the discovered leads before launching expensive experimental studies. We spell out here the positive values, as well as the limitations and possible solutions to address these issues for future improvement. Moreover, any planned investigations involving human subjects should first be approved by the stakeholder’s IRB (Institutional Review Board) who will oversee the safety of the proposed studies and the role of COVID-KG before any experimental studies are conducted. COVID-KG is a tool to enhance biomedical and clinical research; it is not a tool for direct clinical application with human subjects.

\subsection*{Limitations of System Performance and Data Collection} 

\textbf{System errors.} Our system can effectively convert a large amount of scientific papers into knowledge graphs, and can scale as literature volume increases. However, none of our extraction components is perfect, they produce about 6\%-22\% false alarms and misses as reported in section~\ref{sec:kgconstruction}. But as we described in the workflow, all of the connections and answers will be validated by domain experts by checking their corresponding sources before they are included in the drug repurposing report.  COVID-KG is developed for pre-clinical research to target down drugs of interest for biomedical scientists. Therefore, no human subjects or specific populations are directly subjected to COVID-KG unless approved by the stakeholder's IRB who oversees the safety and ethical aspects of the clinical studies in accordance with the Belmont report (https://www.hhs.gov/ohrp/regulations-and-policy/belmont-report/index.html). Accordingly, COVID-KG will not impose direct harm to vulnerable human cohorts or populations, unless misused by the stakeholders without IRB approval. With regards to potential harm in preclinical studies, users of COVID-KG are advised to verify the accuracy of the discovered leads in the source information before conducting expensive experimental studies.

\textbf{Bias in training data.} Proper use of the technology requires that input documents are legally and ethically obtained. Regulation and standards (e.g. GDPR\footnote{The General Data Protection Regulation of the European Union  https://gdpr.eu/what-is-gdpr/.}) provide a legal framework for ensuring that such data is properly used and that any individual whose data is used has the right to request its removal. In the absence of such regulation, society relies on those who apply technology to ensure that data is used in an ethical way. The input data to our system is peer-reviewed publicly available scientific articles. An additional potential harm could come from the output of the system being used in ways that magnify the system errors or bias in its training data. The various components in our system rely on weak distant supervision based on large-scale external knowledge bases and ontologies that cover a wide range of topics in the biomedical domain. Nevertheless, our system output is intended for human interpretation. We do not endorse incorporating the system’s output into an automatic decision-making system without human validation; this fails to meet our recommendations and could yield harmful results. In the cited technical reports for each component in our framework, we have reported detailed error rates for each type of knowledge element from system evaluations and provide detailed qualitative analysis and explanations. 

\textbf{Bias in development data.} We also note that the performance of our system components as reported is based on the specific benchmark datasets, which could be affected by such data biases. Thus questions concerning generalizability and fairness should be carefully considered. Within the research community, addressing data bias requires a combination of new data sources, research that mitigates the impact of bias, and, as done in~\cite{Mitchell2019}, auditing data and models. Sections~\ref{sec:kgconstruction} and \ref{sec:data} cite data sources used for training to support future auditing. A general approach to properly use our system should incorporate ethics considerations as the first-order principles in every step of the system design, maintain a high degree of transparency and interpretability of data, algorithms, models, and functionality throughout the system, make software available as open source for public verification and auditing, and explore countermeasures to protect vulnerable groups. In our ongoing and future work, we have kept increasing the annotated dataset size, add more rounds of user correction and validation, and iteratively incorporate feedback from domain experts who have used the tool, to create new benchmarks for retraining model and conducting more systematic evaluations. 
We recommend caution of using our system output until a more complete expert evaluation has occurred. 

\textbf{Bias in source.} Furthermore, our system output may include some biases from the sources, by way of biases in the peer reviewing process. In our previous work~\cite{yu2014wisdom,ma2015faitcrowd,zhi2015modeling,Zhang2019expertise}, we have aggregated source profile, knowledge graphs and evidence for fact-checking across sources. We plan to extend our framework to include fact-checking to enable practitioners and researchers to access to up-to-the-minute information. 

\textbf{Bias in test queries.} Finally, the queries (i.e., the lists of candidate drugs and proteins/genes) are provide by the users who might have bias in their selection. Addressing the user's own biases falls outside the scope of our project, but as we have stated in the previous subsection, we direct users to carefully examine source information (author, publication date, etc.) and detailed evidence (contextual sentences and documents) associated with the extracted connections.

\section*{Acknowledgement}
This research is based upon work supported in part by U.S. DARPA KAIROS Program No. FA8750-19-2-1004, U.S. DARPA AIDA Program \# FA8750-18-2-0014, .S.  DTRA HDTRA I -16-1-0002/Project \#1553695, eTASC - Empirical Evidence for a Theoretical Approach to Semantic Components, U.S. NSF No. 1741634, the Office of the Director of National Intelligence (ODNI), and Intelligence Advanced Research Projects Activity (IARPA) via contract FA8650-17-C-9116. The views and conclusions contained herein are those of the authors and should not be interpreted as necessarily representing the official policies, either expressed or implied, of DARPA, or the U.S. Government. The U.S. Government is authorized to reproduce and distribute reprints for governmental purposes notwithstanding any copyright annotation therein.

\bibliographystyle{acl_natbib}
\bibliography{emnlp2020,anthology}

\end{document}